\documentclass[letterpaper, 10 pt, journal, twoside]{IEEEtran}

\IEEEoverridecommandlockouts

\usepackage[noend]{algpseudocode}
\usepackage{algorithm}
\usepackage{amsmath,amsfonts}
\usepackage{array}
\usepackage{booktabs}
\usepackage{cite}
\usepackage{graphicx}
\usepackage[colorlinks=true, allcolors=black]{hyperref}
\usepackage{multirow}
\usepackage{orcidlink}
\usepackage[caption=false,font=normalsize,position=bottom]{subfig}
\usepackage{stfloats}
\usepackage{textcomp}
\usepackage{url}
\usepackage{verbatim}
\usepackage{xfrac}

\usepackage{xstring,xifthen}
\newcommand{\subsubsec}[1]{\vspace{0.2em}\noindent\textbf{\IfEndWith{#1}{.}{#1}{#1.}}}
\usepackage{xspace}
\makeatletter
\DeclareRobustCommand\onedot{\futurelet\@let@token\@onedot}
\def\@onedot{\ifx\@let@token.\else.\null\fi\xspace}
 
\def\ie{\emph{i.e}\onedot}

\def\etal{\emph{et al}\onedot}
\makeatother
\DeclareMathAlphabet\mathbfcal{OMS}{cmsy}{b}{n}
\title{\LARGE \bf
Semi-Perspective Decoupled Heatmaps for \\ 3D Robot Pose Estimation from Depth Maps
}

\author{
Alessandro Simoni$^{1}$~\orcidlink{0000-0003-3095-3294}, Stefano Pini$^{1}$~\orcidlink{0000-0002-9821-2014}, Guido Borghi$^{2}$~\orcidlink{0000-0003-2441-7524}, Roberto Vezzani$^{1}$~\orcidlink{0000-0002-1046-6870}
\thanks{Manuscript received: February, 24, 2022; Revised April, 25, 2022; Accepted June, 26, 2022.}
\thanks{$^{1}$Alessandro Simoni, Stefano Pini and Roberto Vezzani are with the Department of Engineering ``Enzo Ferrari'', University of Modena and Reggio Emilia, Italy (\tt \footnotesize alessandro.simoni@unimore.it, s.pini@unimore.it, roberto.vezzani@unimore.it)}
\thanks{$^{2}$Guido Borghi is with the Department  of  Computer  Science  and  En-gineering,  University  of  Bologna,  Italy (\tt \footnotesize guido.borghi@unibo.it)}
}

\begin{document}

\maketitle

\begin{abstract}
Knowing the exact 3D location of workers and robots in a collaborative environment enables several real applications, such as the detection of unsafe situations or the study of mutual interactions for statistical and social purposes. In this paper, we propose a non-invasive and light-invariant framework based on depth devices and deep neural networks to estimate the 3D pose of robots from an external camera. The method can be applied to any robot without requiring hardware access to the internal states. We introduce a novel representation of the predicted pose, namely \textit{Semi-Perspective Decoupled Heatmaps} (SPDH), to accurately compute 3D joint locations in world coordinates adapting efficient deep networks designed for the 2D Human Pose Estimation. The proposed approach, which takes as input a depth representation based on XYZ coordinates, can be trained on synthetic depth data and applied to real-world settings without the need for domain adaptation techniques. To this end, we present the \textit{SimBa} dataset, based on both synthetic and real depth images, and use it for the experimental evaluation.
Results show that the proposed approach, made of a specific depth map representation and the SPDH, overcomes the current state of the art.
\end{abstract}



\section{Introduction}\label{sec:introduction}
Collaborative robots, or \textit{cobots}~\cite{peshkin2001cobot}, have entered the automation market for several years now. They have achieved a rather rapid and wide diffusion, also in the corporate world, thanks to the newly introduced paradigm of interaction~\cite{goodrich2008human} and collaboration \cite{villani2018survey}. About 20 years after their introduction, they still have unexplored potential and challenges that have not yet been fully investigated and solved in the literature.

Among others, the knowledge of the instantaneous pose of robots and humans is a key element to set up an effective and fruitful collaboration between them, allowing several applications, ranging from solutions for the safety of the interaction~\cite{colgate2008safety} to the behavior analysis~\cite{mitsunaga2008adapting}. 

Despite robots usually provide their encoder status through dedicated communication channels, enabling the estimation of their pose and the interaction level~\cite{geravand2013human} through forward kinematics, an external method is desirable in certain cases. For example, the robot controller could be designed by third parties that disable or revoke any permission to access the robot encoders. Another use case is the study of the interaction between collaborative robots and humans, \textit{e.g.} focusing on the human prejudice and distrust against robots~\cite{paulikova2021analysis,palazzi2016spotting}. In this context, a portable and autonomous setup that does not require any hardware access to the internal states of the robots is preferred. In our experience, such a system has been often accepted by manufacturing companies that are participating in an ongoing study.

Therefore, in this paper, we propose to use non-intrusive and ready-to-use sensors, \ie cameras, to address the 3D \textit{Robot Pose Estimation} (RPE) task and, in this way, to monitor the posture of a given robot by means of the position of its joints in world coordinates. Different solutions have been explored to this aim, all of them requiring the unpractical application of specific sensors~\cite{hasegawa2010development} or markers~\cite{kalaitzakis2021fiducial} on the robot structure. Differently from these, we propose to use depth cameras~\cite{sarbolandi2015kinect}, which provide light-invariant and precise 3D scene information at a low cost~\cite{ye2013survey}, and deep neural networks to directly and accurately predict the 3D location of the robotic joints. Since supervised deep networks usually require a great amount of labeled data, we design our approach to leverage synthetic data during the training phase and seamlessly work with real cameras and robots at inference. In this way, our method does not need to acquire data in the real world and annotate them, which is known to be an expensive and time-consuming activity.

To this end, we collected and publicly release\footnote{\url{https://aimagelab.ing.unimore.it/go/rpe}} a new dataset, namely \textit{SimBa}, that contains synthetic data for training and real-world annotated recordings for evaluation. Regarding the model architecture, we draw knowledge and expertise from the related field of the \textit{Human Pose Estimation} (HPE)~\cite{munea2020progress}, being aware of the impressive progress of the computer vision community in that field. Indeed, we present an approach that consists of a novel and effective pose representation, here referred to as \textit{Semi-Perspective Decoupled Heatmaps} (SPDH), that extends the well-known 2D heatmaps to the 3D domain. 
Existing 2D HPE architectures developed for the RGB domain can be easily adapted to process depth data as input and predict the proposed SPDH, leading to accurate 3D joint locations in world coordinates.

We demonstrate that this approach overcomes alternative methods in terms of accuracy, adding negligible computational overhead to existing 2D methods. Experimental results confirm the efficacy of the proposed system, paving the way for future research in the field of the 3D Robot Pose Estimation from depth maps.

To sum up, our main contributions are:
\begin{itemize}
    \item We address the problem of the 3D RPE from depth maps, introducing SPDH, a novel heatmap-based 3D pose representation that can be applied to existing 2D human pose estimation architectures with minimal changes, achieving competitive results.
    
    \item We propose an effective and practical training procedure, based on synthetic depth maps, that can be applied to acquire data at scale without expensive and time-consuming manual annotations. We publicly release the simulation parameters and the dataset used for the experimental validation.
\end{itemize}

\begin{figure*}
    \centering
    \includegraphics[width=\linewidth]{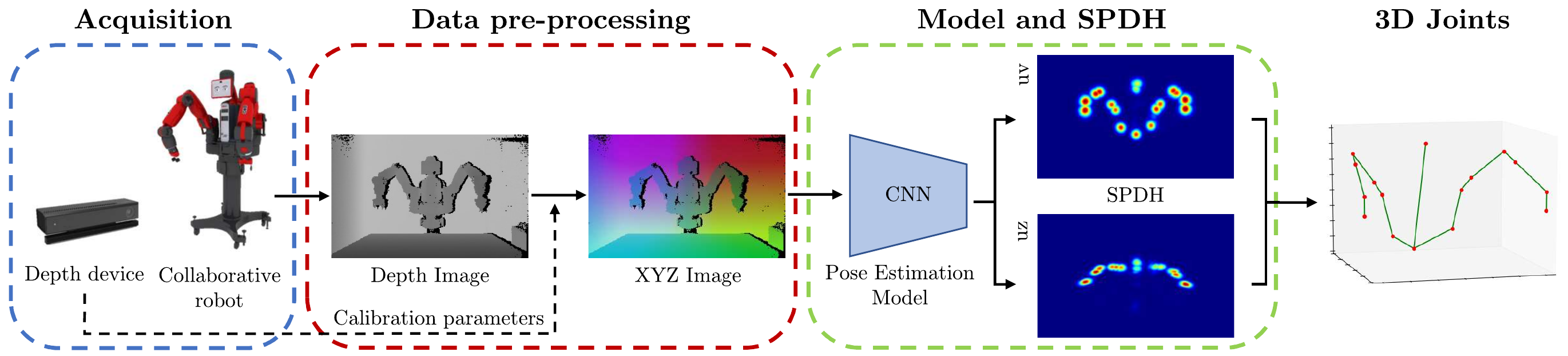}
    \caption{Overview of our 3D Robot Pose Estimation (RPE) system. 
    A depth image is converted into an XYZ image which is given as input to a pose estimation deep model. The network predicts the proposed \textit{Semi-Perspective Decoupled Heatmaps} (SPDH) from which the 3D robot pose is computed.}
    \label{fig:system_overview}
    \vspace{-0.5em}
\end{figure*}


\section{Related work}\label{sec:relatedwork}

\subsection{Robot Pose Estimation (RPE)}
Only a few works address the RPE task and, to the best of our knowledge, only a minor subset of them make use of depth data. This is the case of the system proposed by Bohg \etal~\cite{bohg2014robot}, in which, taking inspiration from~\cite{shotton2012efficient}, a random forest classifier is applied to depth images to segment the links of the robot arms, from which the skeleton joints are estimated. In a similar work~\cite{widmaier2016robot}, Widmaier \etal propose to directly regress the joint angles without the need to predict robot arm segmentation. However, these methods do not estimate the pose in terms of camera-to-robot coordinates.

Instead of depth maps, the large majority of works focus on RGB images.
Labbe \etal~\cite{labbe2021single} proposed a method that, given a single RGB image of a known articulated robot, estimates the 6D camera-to-robot pose in terms of rigid translation and rotation through a render-and-compare approach. A reference point and an anchor part are needed to perform the estimation, and their choice significantly affects the performance. In the work by Lee \etal~\cite{lee2020camera}, the RGB input image is fed to a deep encoder-decoder architecture that outputs one map per keypoint. The final camera-to-robot pose is computed through \textit{Perspective-n-Point} (PnP)~\cite{li2012robust}, assuming that the camera intrinsics and joint configuration of the robot are known. Similarly, in~\cite{lambrecht2019towards} the PnP is used to compute the camera-to-robot pose using a combination of both synthetic and real data. A double system is presented in the work of Tremblay \etal~\cite{tremblay2020indirect}: one network predicts the object-to-camera pose while another estimates the robot-to-camera pose. Both networks are trained entirely on synthetic data and the final output is intended to help the robot grasping system, rather than estimating the whole robot pose. Differently from the discussed approaches, our method can be adapted to any heatmap-based 2D pose estimation method to estimate the 3D pose of a robot from a single depth map.

\subsection{Human Pose Estimation (HPE)}
The task of estimating the human pose has been extensively investigated in the computer vision community.
Similar to the RPE field, the vast majority of works is based on RGB images and outputs only 2D predictions.
Recently, several works also addressed the task of 3D HPE from single monocular intensity images, such as in~\cite{pavlakos2017coarse} where a coarse-to-fine prediction scheme based on volumetric predictions is exploited to compute both the 2D and then the 3D pose. However, these volumetric methods~\cite{ji2020survey} are often characterized by computational inefficiencies, in terms of complexity and memory requirement, even though some recent works~\cite{fabbri2020compressed} attempt to address this issue. These considerations drove our choice to focus on HPE architectures that predict the 2D joint positions through heatmaps applied on RGB images, analyzed in the following.

Among the numerous HPE methods based on 2D heatmaps~\cite{zheng2020deep},
the architecture known as \textit{Stacked Hourglass}, introduced in by Newell \etal~\cite{newell2016stacked}, is developed to process features at different scales and to capture the spatial relationships of the human body, obtaining high accuracy. Recently, Sun \etal~\cite{sun2019deep} proposed a multi-scale approach called High-Resolution Network, or simply \textit{HRNet}. HRNet maintains high-resolution representations throughout the entire estimation process. Other works have been proposed to specifically reduce the computational complexity of the existing methods maintaining a satisfactory accuracy. This is the case of the work described in~\cite{martinez2019efficient}, that proposed the \textit{Fast Pose Machine} (FPM) architecture, based on a cascade of detectors with lightweight and efficient CNN structures. The model can employ different backbones as feature extractors such as \textit{Squeezenet}~\cite{iandola2016squeezenet} and \textit{MobileNet}~\cite{howard2017mobilenets}. Differently from these methods, the simple architecture proposed by Martinez \etal~\cite{martinez2017simple} aims to predict the 3D human pose directly from its 2D version. Despite the simplicity of the approach, the reported results reveal a good accuracy independently from the 2D pose detector used during the training phase. Being aware of the high accuracy achieved by these methods, we aim to adapt RGB HPE architectures for depth data and use them as backbone of our method.

Only a limited number of HPE methods are based on depth maps. In the pioneering work of Shotton \etal~\cite{shotton2011real}, a random forest classifier is used to classify each pixel of the input depth maps and thus to segment the human body. Then, the 3D joint candidates are selected through a weighted density estimator. In~\cite{d2019manual,ballotta2018head} authors propose to use a 2D HPE model to predict head position and human poses in depth images. Schnurer \etal~\cite{schnurer2019real} propose to optimize the Stacked Hourglass~\cite{newell2016stacked} architecture reducing its computational load. The predicted 2D pose is then used in combination with a predicted joint-specific depth map in order to obtain the final 3D coordinates of skeleton joints. In other words, the authors proposed a system that combines a heatmap-based prediction for the 2D coordinates and a value regression for the depth value. A Residual Pose Machines is used by Martinez \etal~\cite{martinez2018real} to detect only the 2D location of human skeleton joints on the depth images. The depth value of the surface close to a given skeleton joint can be computed by sampling from the depth map using the location of that joint. However, this simple approach is not robust against possible body occlusions and the sensor noise; both of them can significantly alter the depth value in the sampled point. Moreover, this approach can only predict the position on the surface of the robot arm, rather than its center. Finally, we observe that the 3D pose estimation from depth data is not yet deeply investigated in the literature, in particular whether deep learning algorithms are used.

\section{Semi-Perspective Decoupled Heatmaps}
\begin{figure}[th!]
    \centering
    \includegraphics[width=1\linewidth]{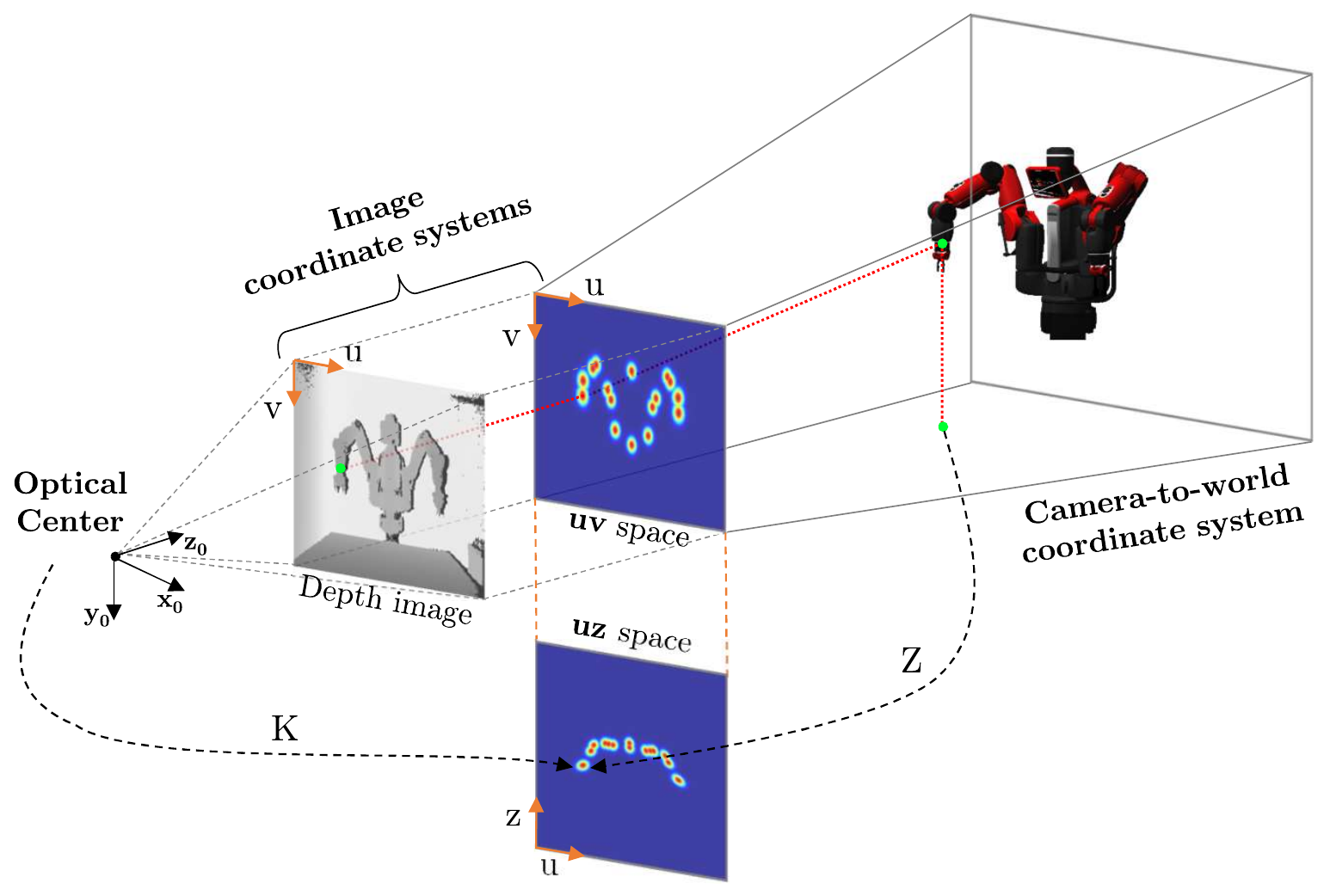}
    \caption{Visual representation of the proposed \textit{Semi-Perspective Decoupled Heatmaps} (SPDH). In particular, $uv$ and $uz$ spaces are depicted in relation to the input depth map and the acquired robot. }
    \label{fig:spdh}
    \vspace{-0.5em}
\end{figure}

We propose a novel \textit{Semi-Perspective Decoupled Heatmaps} (SPDH) pose representation that relies on projections of the 3D space under the assumption of having a single robot in the image. Each 3D joint location is mapped into two decoupled bi-dimensional spaces: the $uv$ space, \ie the camera image plane, and the $uz$ space, composed by quantized $Z$-values and the $u$ dimension, as depicted in Fig.~\ref{fig:spdh}. The pose estimation algorithm will be trained to generate two heatmaps for each joint, one for each space. The corresponding training heatmaps are Gaussian probability distributions centered on the projections of the joint coordinates.

The $uv$ heatmap takes inspiration from the output representation used by most of the recent 2D HPE methods~\cite{zheng2020deep}. However, differently from them, each heatmap is constructed using a Gaussian function that has perspective awareness of the joint's distance from the camera.

Formally, given a joint $j$, the related heatmap $\mathcal{H}^{uv}_j$ is defined in the $uv$ space and computed as follows:
\begin{equation}
    \begin{aligned}
        \mathcal{H}_{j}^{uv} (p) &= \mathcal{N}(p - p_j,\sigma_{j}) \\
        &= \frac{1}{2\pi\sigma_{j}}e^{-[(p^x - p_j^x)^2 + (p^y - p_j^y)^2] / (2\sigma_{j}^2)} \\
        \sigma_{j} &= \frac{\sigma^{m} \cdot f}{Z_{j}}
    \end{aligned}
\end{equation}
where $f$ is the focal length of the camera, $p_j$ is the 2D joint location, $p$ is the pixel location, $Z_j$ is the $Z$ coordinate of the 3D joint location and $\sigma^{m}$ is the desired standard deviation of the Gaussian distribution in the metric space.

On the other hand, the $uz$ heatmap can be seen as the representation of the probability of seeing a joint at the image coordinate $u$ if it were at a distance $z$ from the camera. To generate $\mathcal{H}^{uz}_j$, the following two-step process is applied.

Firstly, we define a restricted depth space $\bar{Z} = \{\bar{Z}_i \in Z;~ \bar{Z}_{\textit{min}} \leq \bar{Z}_i \leq \bar{Z}_{\textit{max}}\}$ and we split it into slices of size $\Delta{Z}$. The number of slices represents the height of the $uz$ heatmap and can be computed as follows:
\begin{equation}
    z = \frac{\bar{Z}_{max} - \bar{Z}_{min}}{\Delta{Z}}    
\end{equation}
In this way, there is a direct connection between the 2D $uz$ space and the depth-aware $\bar{Z}$ space. For simplicity, we sample the space $\bar{Z}$ so that $z$ has the same value of $v$, \ie the dimension of the two heatmaps is the same.

Secondly, we proceed with the computation of the heatmap. We project each point $p^{xy}$ of the image into the 3D reference frame according to its $(x,y)$ coordinates. We apply the intrinsic parameters of the camera $K$,
\ie the focal length $f$ and the optical center $c$, and obtain 
\begin{equation}
    P(p) = \big((p^x - c) \cdot \frac{\bar{Z}^y}{f}, \, \bar{Z}^y\big)
\end{equation}
where $\bar{Z}^y$ is the corresponding value in camera coordinates of $z$ sampled in $p^y$. Then, we compute the euclidean distance $d(p) = ||P(p) - P_j||$ between each point $P$ -- corresponding to each location $p^{xy}$ on the $uz$ space -- and the 3D ground-truth joint location $P_j = [X_j, Z_j]$ -- excluding the $Y$ axis. We use this distance to compute the value of the heatmap in each point $p$ as:
\begin{equation}
    \begin{aligned}
        \mathcal{H}^{uz}_{j}(p) &= \mathcal{N}(P(p) - P_j, \sigma^{m}) \\
        &= \frac{1}{2\pi\sigma^{m}}e^{-d(p) / (2{\sigma^{m}}^2)} \\
    \end{aligned}
\end{equation}
A visual representation of the relation between the proposed Semi-Perspective Decoupled Heatmaps and the 3D space is shown in Fig.~\ref{fig:spdh}; examples of SPHD are also reported in Fig.~\ref{fig:uv-uz-heatmaps}, third and fourth row.

\section{3D Robot Pose Estimation}\label{sec:method}
The proposed approach for the 3D Robot Pose Estimation is based on the aforementioned SPDH  representation. Fig.~\ref{fig:system_overview} depicts a visual overview of each step, detailed in the following subsections, adopted to output the final prediction. 

\subsection{Depth Data Acquisition}
The first step is the data acquisition procedure based on a depth camera, \ie a device capable to acquire depth maps, and a collaborative robot. Despite the challenges and limitations posed by the usage of depth sensors~\cite{pini2021systematic}, we observe that depth data acquired through the same depth device do not usually require the adoption of challenging domain randomization techniques~\cite{tobin2017domain} typically applied on RGB synthetic images to bridge the gap between real and synthetic data~\cite{lee2020camera,tremblay2018training}. In fact, depth data provide robustness to light changes and variations in background textures~\cite{sarbolandi2015kinect}, helping to make the transition from synthetic to real depth data more straightforward and the synthetic data generation easier and less time-consuming.
Moreover, depth cameras provide 3D information of the scene that the method can leverage to estimate the 3D pose of the robot.

Formally, we define an acquired depth map as the couple $D_M = \langle D,K \rangle$, composed by the matrix of distances $D=\{d_{ij}\},\, d_{ij} \in [0, R]$, in which values are between $0$ and the maximum depth range $R$, and $K$, \ie the perspective projection matrix computed with the intrinsic parameters of the acquisition device.
In particular, $d_{ij}$ represents the distance between the optical center and a surface containing the point $p_{ij}$ and parallel to the image plane; then, $D$ can be visualized as a depth image, encoded as one-channel gray-level image $I_\text{D}$, as reported in the first line of Fig.~\ref{fig:uv-uz-heatmaps}.

\subsection{Data pre-processing}
\begin{figure}
    \centering
    \subfloat[\small{(a) Synthetic}]{\includegraphics[width=0.4\linewidth, height=3.8in]{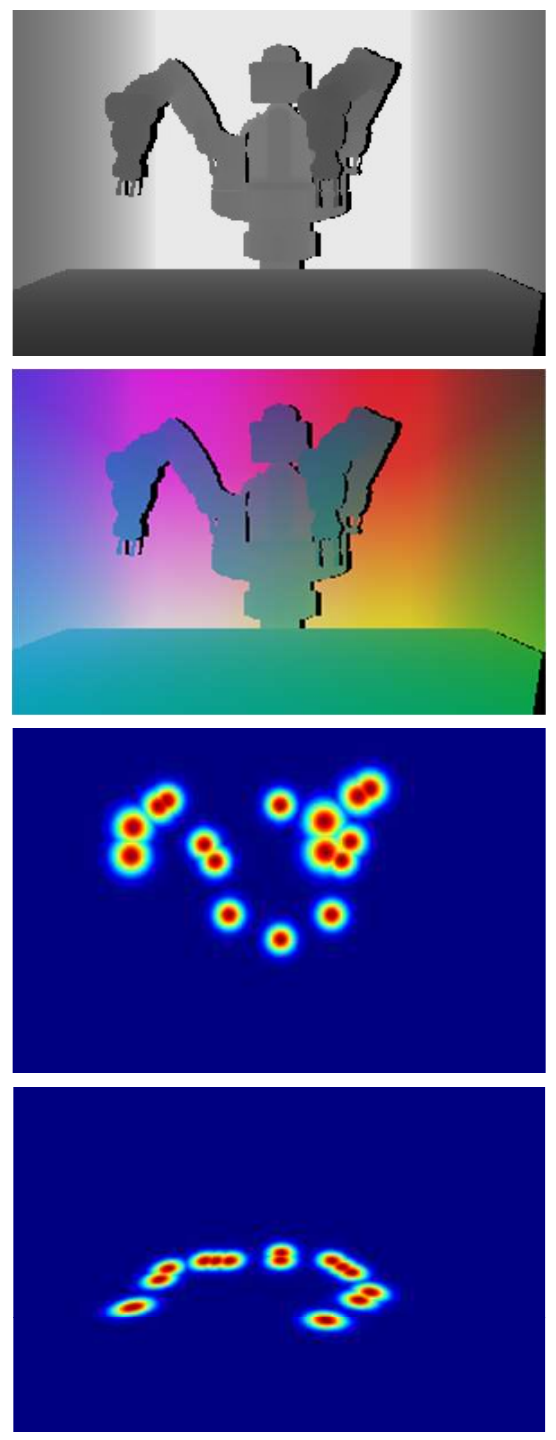}}{}
    \subfloat[\small{(b) Real}]{\includegraphics[width=0.4\linewidth, height=3.8in]{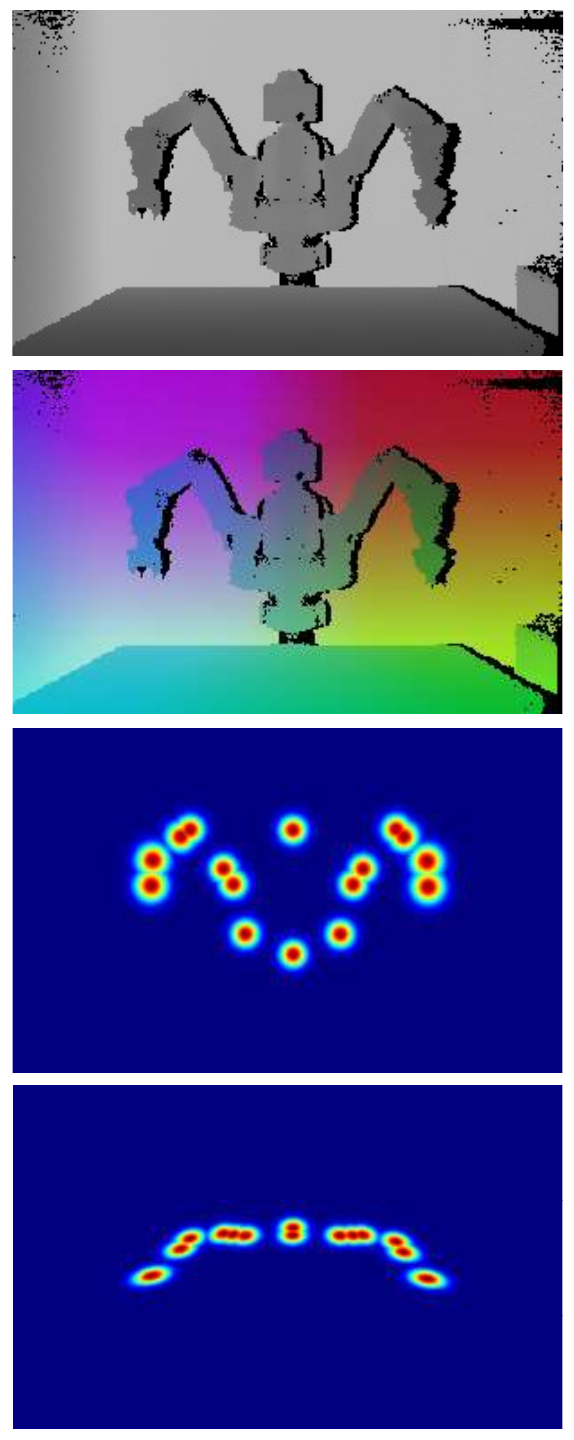}}{}
    \vspace{0.5em}
    \caption{Examples for synthetic and real depth data. 
    Then, XYZ image and the novel SPDH representation, depicted through both the heatmap in \textit{uv} space and heatmap in \textit{uz} space, are reported.}
    \label{fig:uv-uz-heatmaps}
    \vspace{-0.5em}
\end{figure}

As mentioned, the collected depth map $D_M$ contains the information about the distance between the camera and the objects' surfaces of the acquired scene, while our purpose is to have an input with explicit 3D information.
To this end, we convert the depth image $I_\text{D}^{1 \times H \times W}$ into an XYZ image $I^{3 \times H \times W}_{\text{XYZ}}$, where each pixel $q_{ij} \in \mathbb{R}^3$ corresponds to the projection in the 3D space of the original pixel $d_{ij} \in \mathbb{R}^1$, by applying the inverse of the camera intrinsic matrix $K$ and then multiplying by the corresponding depth value. Thus, each pixel of the resulting image represents the 3D coordinate of that pixel in the depth image: visual samples are shown in the second row of Fig.~\ref{fig:uv-uz-heatmaps}. The resulting image is then normalized independently along the three axes $X$, $Y$, $Z$, before being processed by the network.

\subsection{Model Architecture}
As mentioned above, our method is designed to work with a generic deep learning-based architecture belonging to the 2D HPE field. Thanks to the adopted SPDH representation, it is straightforward to adapt the selected architecture for our 3D RPE task, as detailed in the following. The network takes as input an XYZ image of size $3 \times h \times w$ and outputs a $2n \times h \times w$ tensor, where $n$ is the number of joints. The output represents the $uv$ and $uz$ heatmaps for each keypoint, where each pixel value determines the likelihood that a keypoint lies in that position. To compute the predicted joint $\hat{P}_j = [\hat{X}_j, \hat{Y}_j, \hat{Z}_j]$, we exploit the maximum values of the heatmaps; for the $uv$ map, we get the 2D coordinates $\hat{p}_j$ of the Gaussian peak; for the $uz$ map, we consider just the $z$ coordinate of the Gaussian peak which is then converted into a depth metric value as follows:
\begin{equation} \label{eq:Z_conversion}
    \hat{Z}_j = (z \cdot \Delta Z) + \bar{Z}_{\text{min}}
\end{equation}
Finally, $\hat{p}_j$ is projected in the 3D space by applying the inverse of the camera intrinsic matrix $K$ and then multiplying by $\hat{Z}_j$, obtaining the final prediction $\hat{P}_j$. 

\section{Experiments}\label{sec:experiments}

\begin{figure*}
    \centering
    \subfloat{\includegraphics[width=0.043\textwidth]{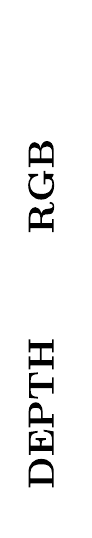}}
    \subfloat[\small{(a) Synthetic}]{\includegraphics[width=0.46\textwidth]{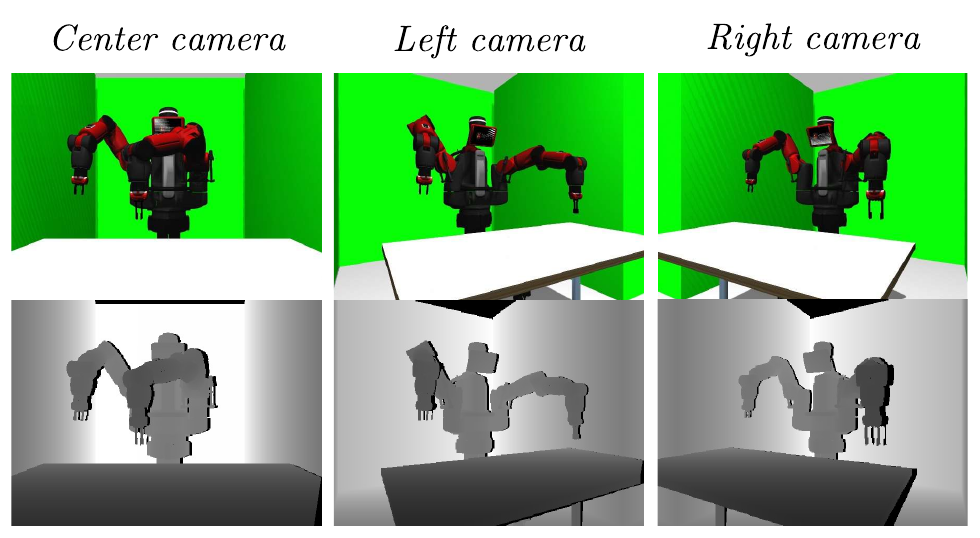}}{}
    \subfloat[\small{(b) Real}]{\includegraphics[width=0.474\textwidth]{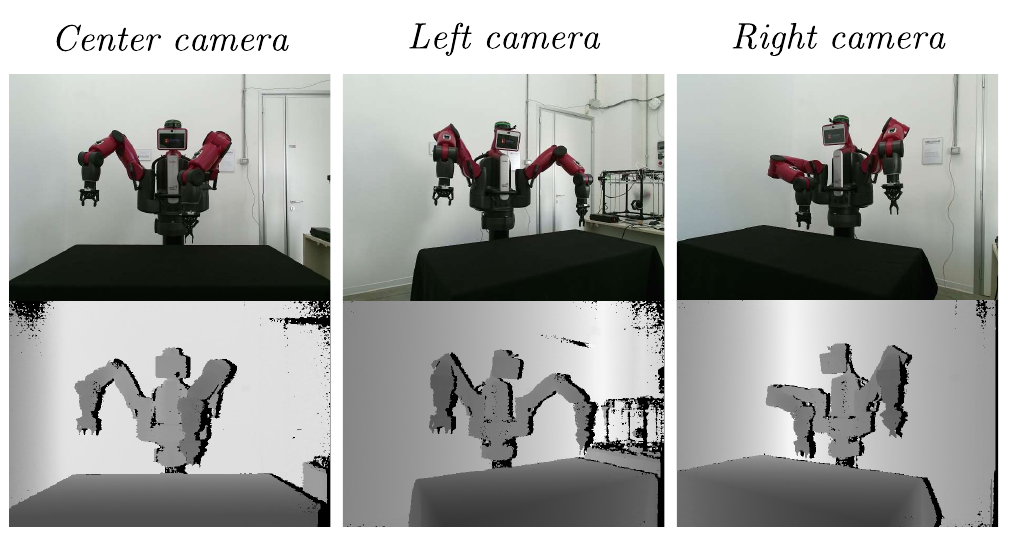}}{}
    \caption{Examples of synthetic and real RGBD frames acquired from different camera positions.}
    \label{fig:dataset_samples}
    \vspace{-0.5em}
\end{figure*}

\subsection{SimBa Dataset}
To evaluate the proposed approach, we collected a new dataset, namely \textit{SimBa}, composed of both synthetic and real images, which are used respectively for training and testing. 

Among several collaborative robots, we choose the \textit{Rethink Baxter}, which has been widely used in the research community. In the collected dataset, the Baxter moves respectively to a set of random pick-n-place locations on a table, assuming realistic poses.

\subsubsec{Synthetic data.} For the synthetic dataset, we use ROS for interacting with the synthetic robot model, the cameras and the environment, and Gazebo for rendering the simulated world. The dataset consists of a set of sequences recorded from $3$ RGB-D cameras (center, left, right) that are randomly moved within a sphere of $1$ meter diameter from their anchor. In particular, we collect two simulation runs with different initializations. Each run contains $20$ recording sequences, composed of $10$ pick-n-place motions which are equally split between the left and right robot arm. Each sequence is recorded at the same time by the three cameras, whose position is randomly changed at the beginning of each sequence. The simulation runs at $10$ fps and records RGB ($1920 \times 1080$) and depth ($512 \times 424$) frames from each camera, the $16$ robot joints positions, the pick-n-place locations, and the camera positions. The synthetic dataset contains a total of 40 sequences with over $350$k annotated RGBD frames.

\subsubsec{Real data.} The real dataset was acquired using the ROS framework. The time-of-flight \textit{Microsoft Kinect One} (second version) was used to record the moving robot from $3$ camera positions (center, left, right). In this case, each camera position contains $20$ pick-n-place sequences which are split equally between the left and right arm. The recording runs at $15$ fps for the RGB ($1920 \times 1080$) and depth ($512 \times 424$) frames and at $40$ fps for the $16$ robot joints positions. The real dataset contains over $20$k annotated RGBD frames.

After the acquisition of the synthetic and the real dataset, we align each depth frame to the RGB sensor using the corresponding extrinsic parameters. Some examples of the frames recorded in both scenarios are depicted in Fig.~\ref{fig:dataset_samples}, in which different levels of precision and noise are visible.

\subsection{Experimental setup}\label{sec:exp_setup}
We split the dataset into $28$ train sequences with over $26$k frames, $4$ validation sequences with over $3.5$k frames, and $8$ test sequences with over $7$k frames. For the training phase, we sampled each synthetic sequence every $10$ frames in order to avoid pose redundancy. The synthetic test set was used only to check the efficacy of the method in the initial phase of the project. To evaluate the method and the competitors, we sampled each sequence of the real dataset every $5$ frames obtaining a test set of $4$k frames.

To generate the ground truth of the joint positions with SPDH representation, we sampled a space $\bar{Z}=[500$ mm$,\, 3380$ mm$]$ with a depth step $\Delta Z = 15$mm. The values of the $\bar{Z}$ space were selected according to the range of the depth sensor that is up to $5$ m~\cite{sarbolandi2015kinect}. Both heatmaps are computed with $\sigma^{m}=50$ mm. Visual examples are depicted in Fig.~\ref{fig:uv-uz-heatmaps}.

We use as input a resized depth image $I_\text{D}$ of resolution $384 \times 192$ applying a 3D data augmentation during training. We first transform the depth image into a pointcloud and then apply a random 3D rotation of $[-5^{\circ}, 5^{\circ}]$ along X or Y axis and a random translation of $[-80\text{mm}, 80\text{mm}]$ along X or Z axis. Then, the pointcloud is converted again into a depth image from which the XYZ image is computed, as explained in Section~\ref{sec:method}.  

We adapt the state-of-the-art 2D HPE architecture called HRNet-32 architecture~\cite{sun2019deep} as the pose estimation model to predict our SPDH from which the 3D robot pose is computed. The network is trained from scratch for $30$ epochs on our synthetic dataset using the $L2$ loss between the predicted and the ground-truth SPDH. We used a batch size of $16$, \textit{Adam}~\cite{kingma2014adam} as optimizer and $1e^{-3}$ as learning rate with $10$ as decay factor after $50\%$ and $75\%$ of the training epochs. At test time, the network is evaluated on the real dataset, without using any domain adaptation techniques, differently from~\cite{lee2020camera} which is based on RGB images.

\begin{table*}
    \centering
    \caption{Comparison between ours and literature approaches trained using an XYZ image as input}
    \begin{tabular}{l l cccc c cc}
    \toprule
        & & \multicolumn{4}{c}{\textbf{mAP} (\%) $\uparrow$} & & \multicolumn{2}{c}{\textbf{ADD} (cm) $\downarrow$} \\
        \cmidrule{3-6} \cmidrule{8-9}
        \textbf{Approach} & \textbf{Network} & $ \mathbf{40}$mm & $\mathbf{60}$mm & $\mathbf{80}$mm & $\mathbf{100}$mm & & \textbf{$L1$} & \textbf{$L2$ } \\
        \midrule
        2D to 3D from depth & Stacked Hourglass (1 HG)~\cite{newell2016stacked} & $8.98$ & $31.21$ & $49.12$ & $66.11$ && $15.63 {\scriptstyle \, \pm 6.62}$ & $11.59 {\scriptstyle \, \pm 5.32}$ \\
        2D to 3D from depth & Stacked Hourglass (2 HG)~\cite{newell2016stacked} & $10.13$ & $31.94$ & $50.54$ & $67.14$ && $14.88 {\scriptstyle \, \pm 6.10}$ & $11.06 {\scriptstyle \, \pm 5.04}$ \\
        2D to 3D from depth & FPM (MobileNet)~\cite{martinez2019efficient} & $9.83$ & $29.09$ & $49.13$ & $66.70$ && $16.25 {\scriptstyle \, \pm 6.66}$ & $11.66 {\scriptstyle \, \pm 5.38}$ \\
        2D to 3D from depth & FPM (SqueezeNet)~\cite{martinez2019efficient} & $10.84$ & $32.87$ & $51.58$ & $67.87$ && $15.12 {\scriptstyle \, \pm 6.11}$ & $11.22 {\scriptstyle \, \pm 5.07}$ \\
        2D to 3D from depth & HRNet-32~\cite{sun2019deep} & $12.52$ & $33.23$ & $49.57$ & $67.18$ && $14.51 {\scriptstyle \, \pm 5.59}$ & $10.86 {\scriptstyle \, \pm 4.64}$ \\
        2D to 3D from depth & HRNet-48~\cite{sun2019deep} & $12.15$ & $32.55$ & $50.83$ & $67.99$ && $14.62 {\scriptstyle \, \pm 5.78}$ & $10.99 {\scriptstyle \, \pm 4.81}$ \\
        \midrule
        3D regression & ResNet-18~\cite{he2016deep} & $9.40$ & $19.99$ & $27.06$ & $44.44$ && $17.10 {\scriptstyle \, \pm 5.43}$ & $12.20 {\scriptstyle \, \pm 4.12}$ \\
        \midrule
        2D to 3D lifting & Martinez et al.~\cite{martinez2017simple}~$^{*}$ & $26.96$ & $37.98$ & $48.40$ & $58.33$ && $14.01 {\scriptstyle \, \pm 4.84}$ & $10.03 {\scriptstyle \, \pm 3.53}$ \\
        \midrule
        Volumetric heatmaps & Pavlakos et al.~\cite{pavlakos2017coarse} & $18.15$ & $42.24$ & $61.60$ & $86.15$ && $10.35 {\scriptstyle \, \pm 1.07}$ & $7.11 {\scriptstyle \, \pm 0.65}$ \\
        \midrule
        \textit{SPDH (ours)} & HRNet-32~\cite{sun2019deep} & $\mathbf{53.75}$ & $\mathbf{79.75}$ & $\mathbf{93.90}$ & $\mathbf{98.12}$ && $\mathbf{6.62} {\scriptstyle \, \pm 1.53}$ & $\mathbf{4.41} {\scriptstyle \, \pm 1.09}$ \\
        \bottomrule
        &* relative joint positions
    \end{tabular}
    \label{tab:competitor}
\end{table*}

\begin{figure*}
    \centering
    \begin{tabular}{ccc}
    \hspace{-1.1em}
    \includegraphics[width=0.329\linewidth]{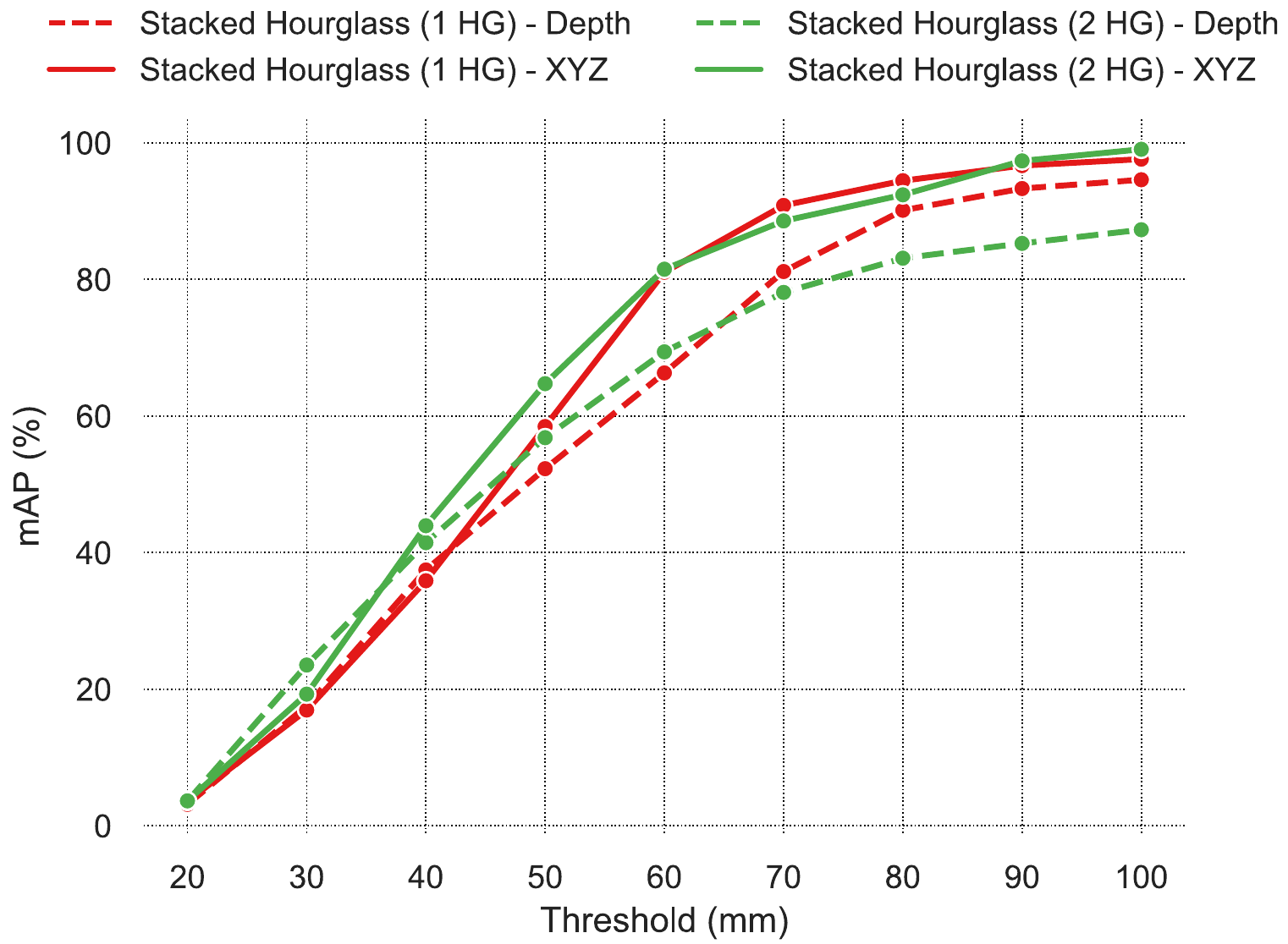}
    \includegraphics[width=0.329\linewidth]{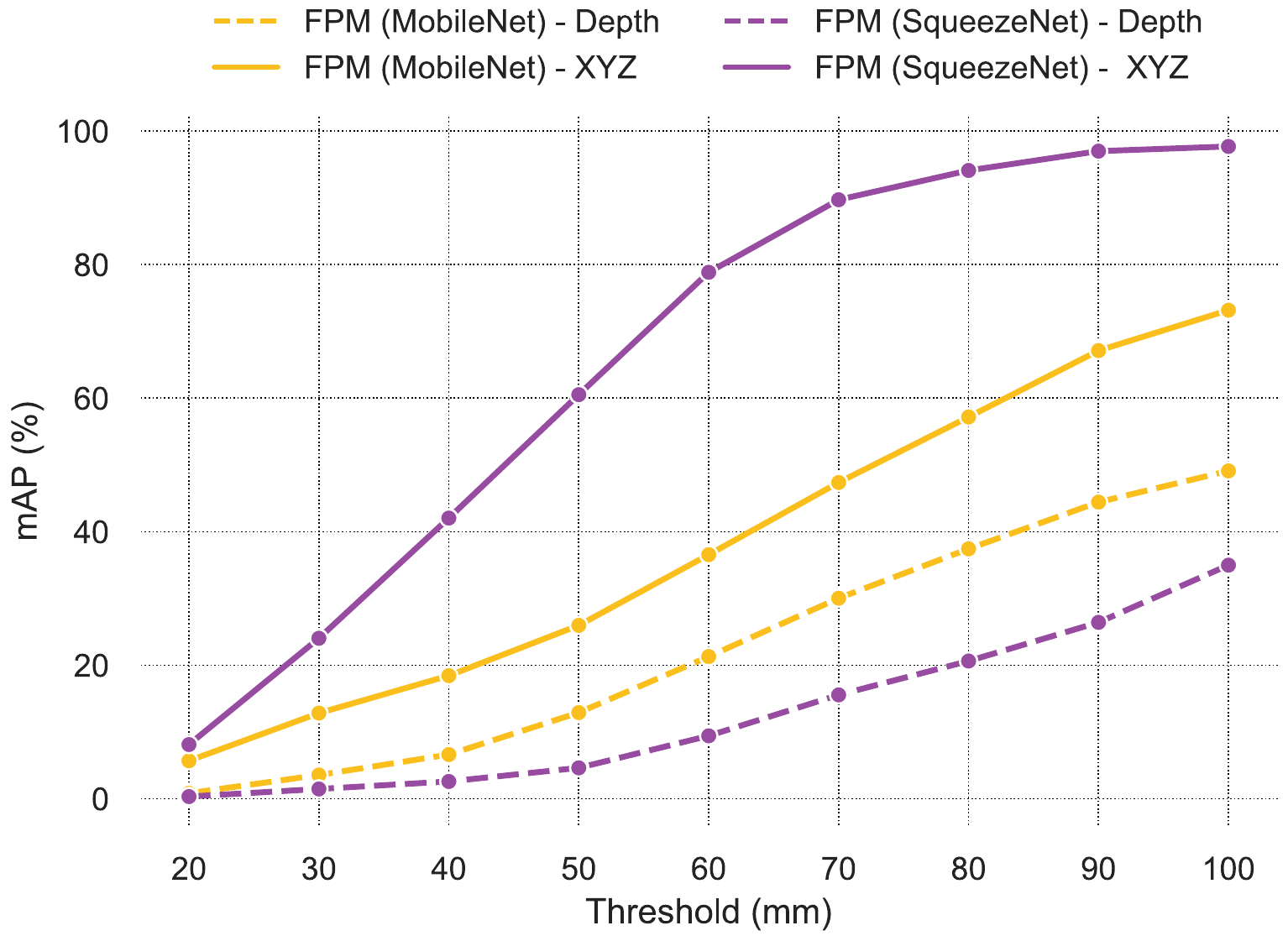}
    \includegraphics[width=0.329\linewidth]{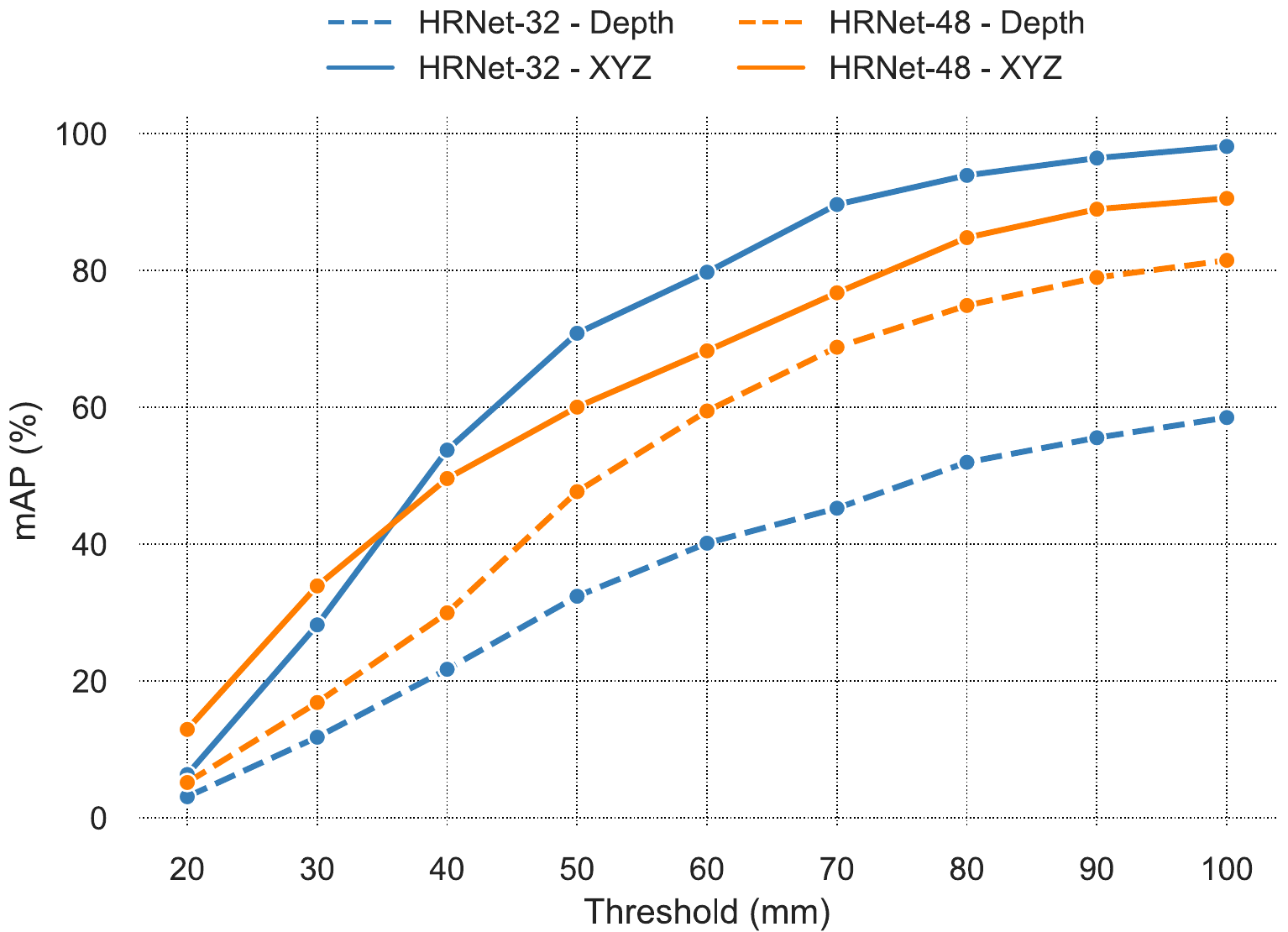}
    \end{tabular}
    \caption{Comparison on mAP scores between different networks trained giving as input just a depth image $I_\text{D}$ or the proposed XYZ image $I_\text{XYZ}$ (dashed line = input $I_\text{D}$, solid line = input $I_\text{XYZ}$, same color = same network configuration).}
    \label{fig:input_comparison}
    \vspace{-0.5em}
\end{figure*}

\subsection{Metrics}
For the 3D evaluation of the predicted robot poses, we exploit the average distance metric (ADD)~\cite{lee2020camera,xiangposecnn}, in terms of the average distance of all 3D robot joints to their ground truth positions. ADD metric is useful in order to merge translation and rotation errors in a single value.
In particular, we compute ADD reporting $L1$ and $L2$ average distances expressed in centimeters with standard deviations w.r.t. ground truth positions. Here, lower results represent good performance. In addition, we compute the \textit{mean Average Precision} (mAP), which expresses the percentage of 3D keypoints within a certain threshold. In our experimental validation, we set $4$ different thresholds at $40, 60, 80, 100$ mm. Here, higher results are better. We believe this metric shows the performance in a more straightforward manner compared to ADD, highlighting the accuracy of the system at different thresholds.

\subsection{Competitors}
In order to validate the proposed RPE approach, we compare it with four alternatives, belonging to the HPE domain, that can be adapted to predict the 3D robot pose:
\begin{itemize}
    \item ``$2$D to 3D from depth'' is a two-step approach. Firstly, a state-of-the-art HPE method~\cite{newell2016stacked, martinez2019efficient, sun2019deep} predicts the 2D robot pose on depth images. Then, the $Z$ value is sampled from the depth to obtain the 3D joint coordinates.
    
    \item the ``$3$D regression'' method corresponds to an architecture that directly regresses the 3D joint coordinates starting from a depth image. We empirically found that the best performance is obtained using the well-known ResNet~\cite{he2016deep} architecture, adapted and trained for regressing the 3D robot joints.
    
    \item the ``$2$D to 3D lifting'' approach directly converts a set of already predicted 2D joint locations to their 3D counterpart, relative to a root joint. In particular, we select the network proposed by Martinez \etal~\cite{martinez2017simple}.
    
    \item a ``volumetric heatmap'' approach that outputs 3D heatmaps. In our experimental validation, we adopt the state-of-the-art method proposed in~\cite{pavlakos2017coarse}, which predicts a volume with size $d \times w \times h$ -- with $d=64$ -- for each joint and uses its maximum value as the 3D joint location.
\end{itemize}

\subsection{Results}
As shown in Table~\ref{tab:competitor}, our method performs better than other competitors in all the metrics,
especially in terms of mAP with low distance thresholds. We observe that the 2D to 3D from depth approach leverages the high precision of 2D pose estimation models, but is limited by the depth map; indeed, it samples the depth values at the 2D joint coordinate, resulting in predicting a 3D location on the robot surface rather than onto the inner joint position. The approach based on direct 3D regression does not reach satisfactory results, confirming that the task is not trivial. The 2D to 3D lifting method uses a relative joints' position with regard to a specific root joint -- the robot base in our case. Thus, this approach is not directly comparable to our proposal, since it needs a post-processing step in which the camera pose has to be known or predicted and then applied to the 3D coordinates in order to get the correct camera-to-robot values. However, we noticed overfitting phenomena on synthetic training data, resulting in low accuracy when testing on real data.

The approach based on volumetric heatmaps obtains a good level of accuracy, even if still lower than our method.
However, the main issue of this approach is the high computational load that predicting 3D volumes requires. Indeed, a volumetric heatmap represents a quantized 3D space that quickly grows in size in order to increase the precision of the method. This problem can be noticed during training when this approach requires almost double the amount of GPU memory ($8.3$GB) than our method ($4.7$GB). Summarizing, our approach achieves the best results in predicting the inner joints of the robot, which is a challenging task to solve with the alternative approaches.

\begin{figure*}[th!]
    \centering
    \includegraphics[width=1\linewidth]{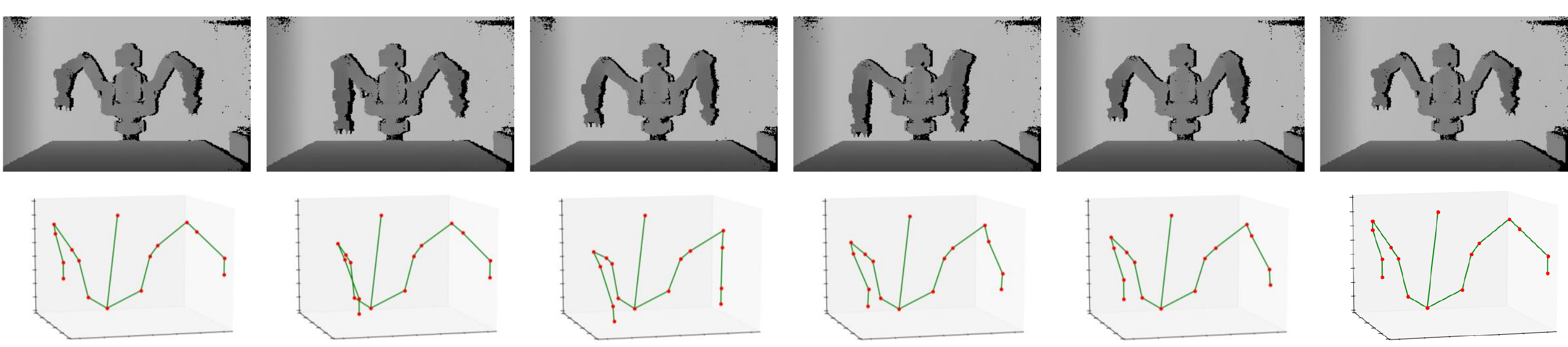}
    \caption{Real depth images and final predicted 3D robot pose for a random pick-and-place motion with both arms.}
    \label{fig:qualitative_results}
    \vspace{-0.5em}
\end{figure*}

Finally, some qualitative results of our method are depicted in Fig.~\ref{fig:qualitative_results}, in which is reported the initial depth image and the final 3D robot skeleton.

\begin{table}[t]
    \centering
    \caption{Comparison of results using different values of $\Delta Z$}
    \resizebox{1\linewidth}{!}{
    \begin{tabular}{l c cccc}
        \toprule
        & & \multicolumn{4}{c}{\textbf{mAP} (\%) $\uparrow$} \\
        \cmidrule{3-6}
        \textbf{Network} & $\mathbf{\Delta Z}$ (mm) & \textbf{40mm} & \textbf{60mm} & \textbf{80mm} & \textbf{100mm} \\
        \midrule
        Stacked HG (2 HG)~\cite{newell2016stacked} & $7.5$ & $46.83$ & $78.08$ & $89.52$ & $96.07$ \\
        Stacked HG (2 HG)~\cite{newell2016stacked} & $15$ & $43.94$ & $81.50$ & $92.39$ & $99.06$ \\
        Stacked HG (2 HG)~\cite{newell2016stacked} & $30$ & $44.55$ & $76.04$ & $87.22$ & $99.03$ \\
        \midrule
        HRNet-32~\cite{sun2019deep} & $7.5$ & $51.80$ & $69.42$ & $79.27$ & $88.24$ \\
        HRNet-32~\cite{sun2019deep} & $15$ & $53.75$ & $79.75$ & $93.90$ & $98.12$ \\
        HRNet-32~\cite{sun2019deep} & $30$ & $43.66$ & $69.24$ & $86.38$ & $97.20$ \\
        \bottomrule
    \end{tabular}
    }
    \label{tab:delta_z_comparison}
    \vspace{-0.5em}
\end{table}

\subsection{Ablation study} \label{sec:ablation}
To further evaluate our approach, we perform an ablation study to investigate the impact of using different network inputs and of sampling a different $\bar{Z}$ space when computing the $uz$ heatmap. 

In the first experiment, we adapt three different 2D HPE baselines~\cite{newell2016stacked, martinez2019efficient, sun2019deep} in order to learn the proposed SPDH output. In particular, they were trained using two different inputs: a depth image $I_\text{D}$ and the proposed XYZ image $I_\text{XYZ}$. As can be seen in Fig.~\ref{fig:input_comparison}, the results confirm that the network learning process benefits from an input $I_\text{XYZ}$ with explicit 3D information. Indeed, using only the depth values from $I_\text{D}$ makes the network independent from the intrinsic parameters of the camera, which are needed to learn a meaningful 3D representation of the world from a 2D space.

With the hypothesis that a smaller value of $\Delta Z$ should take to higher mAP scores, the latter experiment explores different sampling of the $\bar{Z}$ space. We select the networks~\cite{newell2016stacked, sun2019deep} with the best performances and train them with three different $\Delta Z$ values, \ie $7.5$, $15$ and $30$ mm.
We observe that using $\Delta Z = 30$ a bigger space $\bar{Z} = [0, 5760]$mm is sampled, while for $\Delta Z = 7.5$mm we adapt the input for our system increasing the size of input images to $384 \times 384$ adding a upper-lower padding and maintaining the same $\bar{Z}$ space as our main experiment in Section~\ref{sec:exp_setup}. Experimental results reported in Table~\ref{tab:delta_z_comparison} reveal that the choice of the parameter $\Delta Z$, that can be potentially non-trivial since it changes the size of the $uz$ heatmap, does not have a significant impact on the performance of the whole proposed system, tending to avoid the need to ad hoc decrease or increase $\Delta Z$ value for different application contexts.

\section{Limitations and Future Work}
Although our experimental section show promising results, we observe that the output is limited to a single robot in the acquired scene. In addition, the influence of other objects in the scene on the final prediction needs to be investigated, since these objects could change the visual appearance of the scene or produce occlusions on robot surfaces. The proposed system is one of the first attempts in this field, and it can be improved in many terms, \textit{e.g.} on the temporal smoothness of the pose. Indeed, as future work we aim to insert the temporal consistency in the learning process to refine the predicted 3D pose. In this way, the collected dataset can be directly exploited, since it contains video sequences of robotic actions. Moreover, the proposed method can be adapted to predict also the pose of different agents in the scene, including humans, enabling the reasoning on the HRI task through 3D poses. This requires the collection of new synthetic and real data with humans, and the non-trivial 3D annotations of them.

\section{Conclusion}
In this paper, we present a depth-based 3D Robot Pose Estimation approach that can be trained on fully synthetic data and evaluated on real data with promising results. Leveraging from a novel heatmap-based output representation, namely \textit{Semi-Perspective Decoupled Heatmaps} (SPDH), the proposed method takes an XYZ image obtained from a depth map as input and predicts two bi-dimensional heatmaps that are then converted to 3D joint locations. We also present and publicly release the \textit{SimBa} dataset, that we use to evaluate the proposed system in both synthetic and real environments. A thorough experimental section compares the proposed method to alternative approaches derived from the HPE domain, confirming its promising performance.

\section*{Acknowledgment}
The authors would like to thank Margherita Peruzzini and Riccardo Karim Khamaisi of XiLab Unimore for their invaluable help in collecting real sequences with the Baxter robot. The authors are also grateful to Stan Birchfield and Timothy E. Lee of NVIDIA for helpful discussions.

\newpage

\bibliographystyle{IEEEtran}
\bibliography{biblio}

\end{document}